\documentclass[lettersize,journal]{IEEEtran}
\usepackage{amsmath,amsfonts,amssymb}
\usepackage{algorithm}
\usepackage{algpseudocode} 
\usepackage{epsfig}
\usepackage{graphicx}
\usepackage{soul}

\usepackage{paralist}
\usepackage{hyperref}

\usepackage{array}
\usepackage{textcomp}
\usepackage{stfloats}
\usepackage{url}
\usepackage{verbatim}
\usepackage{graphicx}
\usepackage{cite}
\usepackage{tabularx}
\usepackage[dvipsnames]{xcolor}
\usepackage{caption}
\usepackage{subcaption}
\usepackage[normalem]{ulem}
\usepackage{multirow}
\usepackage{cellspace}
\usepackage{arydshln}
\usepackage{booktabs}
\usepackage{times}
\usepackage{array}
\usepackage{float}
\begin{document}

\title{Cross-Spectral Attention for Unsupervised RGB-IR Face Verification and Person Re-identification}
\author{Kshitij Nikhal,~\IEEEmembership{Graduate Student Member,~IEEE,}
    Cedric Nimpa Fondje,~\IEEEmembership{Graduate Student Member,~IEEE,}
        Benjamin S. Riggan,~\IEEEmembership{Member,~IEEE}
\thanks{The authors have no conflicts of interest to declare that are relevant to the content of this article.}
}


\maketitle

\begin{abstract}
Cross-spectral biometrics, such as matching imagery of faces or persons from visible (RGB) and infrared (IR) bands, 
have rapidly advanced over the last decade due to increasing sensitivity, size, quality, and ubiquity of IR focal plane arrays and enhanced analytics beyond the visible spectrum. Current techniques for mitigating large spectral disparities between RGB and IR imagery often include learning a discriminative common subspace by exploiting precisely curated data acquired from multiple spectra. Although there are challenges with determining robust architectures for extracting common information, a critical limitation for supervised methods is poor scalability in terms of acquiring labeled data.  Therefore, we propose a novel unsupervised cross-spectral framework that combines (1) a new pseudo triplet loss with cross-spectral voting, (2) a new cross-spectral attention network leveraging multiple subspaces, and (3) structured sparsity to perform more discriminative cross-spectral clustering.  
We extensively compare our proposed RGB-IR biometric learning framework (and its individual components) with recent and previous state-of-the-art models on two challenging benchmark datasets: DEVCOM Army
Research Laboratory Visible-Thermal Face Dataset (ARL-VTF) and RegDB person re-identification dataset, and, in some cases, achieve performance superior to completely supervised methods.
\end{abstract}


\section{Introduction}
\label{sec:intro}

Biometric recognition involves automatic identification or verification of individuals according to physical characteristics (or modalities), such as face, fingerprint, and iris, that are considered to be common (universal), discriminative (unique), stationary (permanent), and measurable (collectable) among large populations~\cite{biometric_intro}.  Given the potential to acquire identifiable information from extended standoff distances, face and person (whole body) modalities are more conducive to enhancing surveillance capabilities in various operational scenarios.
However, our capacity to label and utilize large amounts of data for machine learning is constrained by the time and costs associated with the annotation process. This is notably apparent in defense and intelligence missions, which involves cross-spectral conditions (Infrared (IR) to Visible (RGB) matching) for nighttime recognition.

\begin{figure}[tb]
     \centering
     
         \includegraphics[width=\columnwidth]{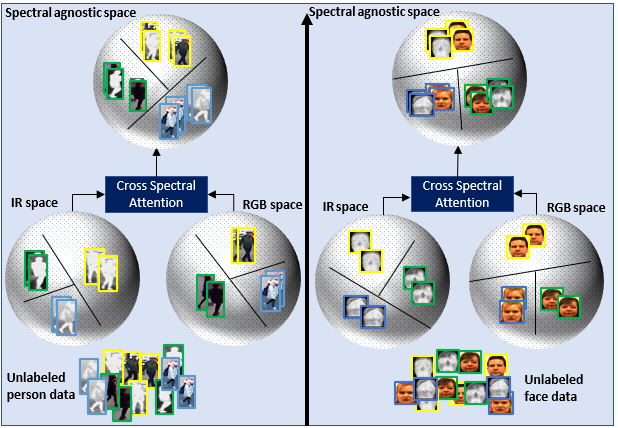}
     \caption{Our proposed unsupervised cross-spectrum framework, which learns spectral invariance via cross-spectral attention without labeled data, applied to ReID (left) and FaceVeri (right). }
     \label{fig:introduction}
\end{figure}

Therefore, we focus on facilitating both `soft' and `hard' biometric tasks---person Re-IDentification (ReID) and Face Verification (FaceVeri), respectively---under challenging unsupervised and cross-spectrum conditions (Fig.~\ref{fig:introduction}).  



ReID is the retrieval of instances of persons that correspond to the same identity from query images or videos captured from multiple cameras and camera views. Its effectiveness relies on associating visually similar personal characteristics with identity information. Despite the plethora of unique characteristics of a person, such as anthropometrics and biometrics, 
ReID is often considered a soft-biometric since visual characteristics like clothing, hair style, footwear, body shape, and gait are used.  These characteristics are not necessarily unique to the individual, but can be combined to build stronger evidences of unique identities \cite{reid2013softbiometrics}.  However, the advantage of person ReID compared to other biometric modalities (e.g., face) is the ability to extract the underlying visual characteristics using low-resolution imagery (i.e., few pixels on persons).


Conversely, FaceVeri relies on the comparison of various facial features, including inter-pupillary distance, nose shape, face type/shape (e.g., round, oval, rectangular, and heart), eyebrow length/thickness, and more, which exhibit a strong correlation with identity information. 
Unlike ReID, which relies on soft-biometric information like clothing and gait, FaceVeri is considered a hard-biometric as it uses more unique and intrinsic personal characteristics.
Significant advancements have been made to extract facial features under challenging settings, including low resolution, pose, and occlusion.  

Despite recent advances in ReID and FaceVeri, persisting challenges include (1) unsupervised learning and (2) cross-spectral matching.  While the majority of ReID and FaceVeri methods exploit supervised learning for RGB biometric tasks, there have been increasing interests in unsupervised learning (or clustering) methods and in cross-spectral learning (e.g., IR to RGB). This interest stems from the ability to utilize large-scale multi-spectral datasets while minimizing the time, cost, and effort needed for meticulous manual data curation.

However, these two methodologies have largely been considered independently for unsupervised learning, since the ability to match faces or persons from different spectra requires strong supervision in the form of labeled data. Therefore, we propose a new framework for unsupervised cross-spectral biometric learning, specifically for ReID and FaceVeri. 

Relevant approaches to unsupervised cross-spectral biometrics include unsupervised domain adaptation techniques, such as cycle Generative Adversarial Network (cycleGAN) \cite{zhu2017unpaired}, Contrastive Adaptation Networks (CANs)~\cite{kang2019contrastiveadapt}, or Maximum Mean Discrepancy (MMD)~\cite{tolstikhin2016mmd} based methods. Formally, unsupervised domain adaptation is characterized by the process of learning from labeled data in the source domain and unlabeled data in the target domain, as defined by ~\cite{pan2009survey}. This approach relies on the underlying assumption that source domain datasets tend to be larger and more frequently labeled than their target domain counterparts.  Despite the advances in unsupervised domain adaptation, these technique are commonly applied when (1) there is a relatively small domain gap between source and target distributions (e.g., Office-31~\cite{office31}, VisDA-2017~\cite{visda2017}, sketch-to-photo~\cite{sketchtophoto}) and (2) there are many exemplars per class caption in classifiers that are exploited during inference. 

While cycleGANs are well-adapted for high-level stylistic domain transforms, they are not inherently designed to encode fine-grained discriminative information to perform cross-spectral matching.  Similarly, MMD~\cite{long2015learning} and Joint MMD (JMMD) largely address the mean shift between domains and are more applicable when domain gaps are relatively small.

Unlike CAN, which leverages labels from the source domain, without loss of generality, we do not use any labels from either domain (source or target) to alleviate potential for over-training on the source data.


To this end, we propose a more discriminative unsupervised domain adaptation framework for cross-spectral ReID and FaceVeri that does not leverage labels from either domain (RGB or IR).
The goal of this paper is to find a domain invariant space that is discriminative among identities. To achieve this, we leverage knowledge from IR and RGB specific subspaces to refine a common space (Fig.~\ref{fig:introduction}). 
RGB imagery contains color and texture information of the person, while thermal IR (wavelengths in the range 7-14 $\mu m$) imagery consist of shape and heat signatures that are invariant to external illumination. Utilizing common and mutually exclusive information from both domains allows the network to be robust to changes that does not contribute to identity discriminability.  

To achieve this, we first implement agglomerative clustering to get reliable intra-domain clusters, similar to BUC~\cite{lin2019bottom} and GAM~\cite{nikhal2021unsupervised}. Agglomerative clustering is a bottom-up approach where each data point is initially treated as a single cluster, and progressively merged together until it forms desired numbers of clusters. Next, we use a novel voting scheme incorporated in the triplet loss, to associate corresponding RGB and IR clusters. Finally, we see further performance improvement by incorporating structured sparsity on the learned representation. 
We train this framework in an end-to-end manner, and  outperform many state-of-the-art (SOTA) supervised methods, setting a strong baseline for unsupervised performance.  

Our primary contributions include: (1) a pseudo triplet loss with a cross-spectral voting scheme sampling, (2) a cross-spectral attention network using augmented subspaces, (3) pixel- and channel-wise structured sparsity representation learning, and (4) extensive analysis on both biometric tasks using the RegDB and ARL-VTF datasets, setting new benchmarks for unsupervised biometric learning.

\section{Related Work}
\label{sec:relatedwork}
While single spectrum FaceVeri and ReID is sufficiently challenging due to varying pose, illumination, and resolution across cameras, cross-spectral FaceVeri and ReID are even more challenging because of large domain gap between the spectra. Our work focuses on solving this dual-spectral problem in a completely unsupervised manner.  

\textbf{RGB Unsupervised ReID:} Multiple approaches have been taken to solve the problem of ReID without using labels.
D-MMD~\cite{DMMD} uses a labeled source dataset to transfer knowledge to an unlabeled target dataset by aligning the target and source pair-wise distance distributions.
EUG~\cite{EUG} uses a single labeled tracklet to assign unlabeled tracklets to its nearest labeled neighbour.
Methods like GAM~\cite{nikhal2021unsupervised} and BUC~\cite{lin2019bottom} use agglomerative clustering to generate pseudo-labels and iteratively optimize the network using a memory bank. 
HDCRL~\cite{cheng2022hybrid} treats unclustered instances (or outliers) as independent classes, and uses contrastive learning to fully utilize the dataset. 
While there exists very limited research in cross-spectral unsupervised ReID, we use the principles of RGB clustering from GAM to get our initial intra-domain clusters. 

\textbf{Cross-Spectral ReID:}
Many supervised cross-spectral ReID methods employ a two stream network, that learn domain-specific and domain-shared features. 
In \cite{xmodality} a lightweight generator is used to translate visible images to an intermediate domain with a domain gap constraint. 
In \cite{learningbualigning} a dense pixel-wise correspondence is learned between RGB and IR images, suppressing spectra-related features. 
CmGAN~\cite{cmgan} trains a generative network with inter-domain triplet loss supervision to generate a cross-spectral representation. 
In~\cite{cheng2023cross}, the model extracts modality-shared
image features and modality discrepancy is eliminated.
H2H~\cite{homhet} uses a labeled Market1501 dataset to generate pseudo-labels for both domains for training, which might be viewed as unsupervised domain adaptation. ADCA~\cite{yang2022augmented} relies on the performance of DBSCAN algorithm to generate pseudo-labels within each domain. 
However, previous work have not explored completely unsupervised learning (generating pseudo-labels through learning) in the cross-domain (or cross-spectral) setting. This work addresses this problem, and achieves superior performance compared to many supervised methods.

\textbf{RGB Unsupervised Face Verification:}
Unsupervised FaceVeri has recently garnered attention due to the high cost associated with annotating large amounts of data. A CNN-based auto-encoder is trained by tracking a face across multiple frames to generate a dataset of positive and negative face examples~\cite{facerepresentation}. In~\cite{solomon2022uface}, the k-most similar and dissimilar images of the input face image is identified to train an auto-encoder for reconstruction. A conditional generative network is trained using constrastive learning strategies to generate synthetic data~\cite{ugan}. In our work, we introduce an unsupervised method for generating pseudo-labels that is capable to work across spectra. 

\textbf{Cross-spectral Face Verification:}
Most works in FaceVeri employ GAN-based methods to synthesize imagery from different spectra.
Pix2Pix~\cite{pix2pix2017} uses conditional adversarial networks to translate images from IR to RGB using a U-Net based architecture. 
GANVFS~\cite{ganvfs} jointly estimates the RGB features and RGB image reconstruction from thermal images using identity and perceptual objectives, to retain discriminative face characteristics. 
SAGAN~\cite{sagan} utilizes a self-attention module to capture long-range dependency information with cycle consistency and a patch discriminator for inter-domain synthesis. 
Other related work include RST~\cite{rst}, where a residual spectral transform is learned to produce domain invariant representations.
In contrast, our work encourages knowledge transfer from both domains as each contributes some information~\cite{sagan} of the discriminative characteristics of the face.

\section{Methodology}
\label{sec:methodology}
\begin{figure*}[tb]
    \centering
\includegraphics[width=1.8\columnwidth]{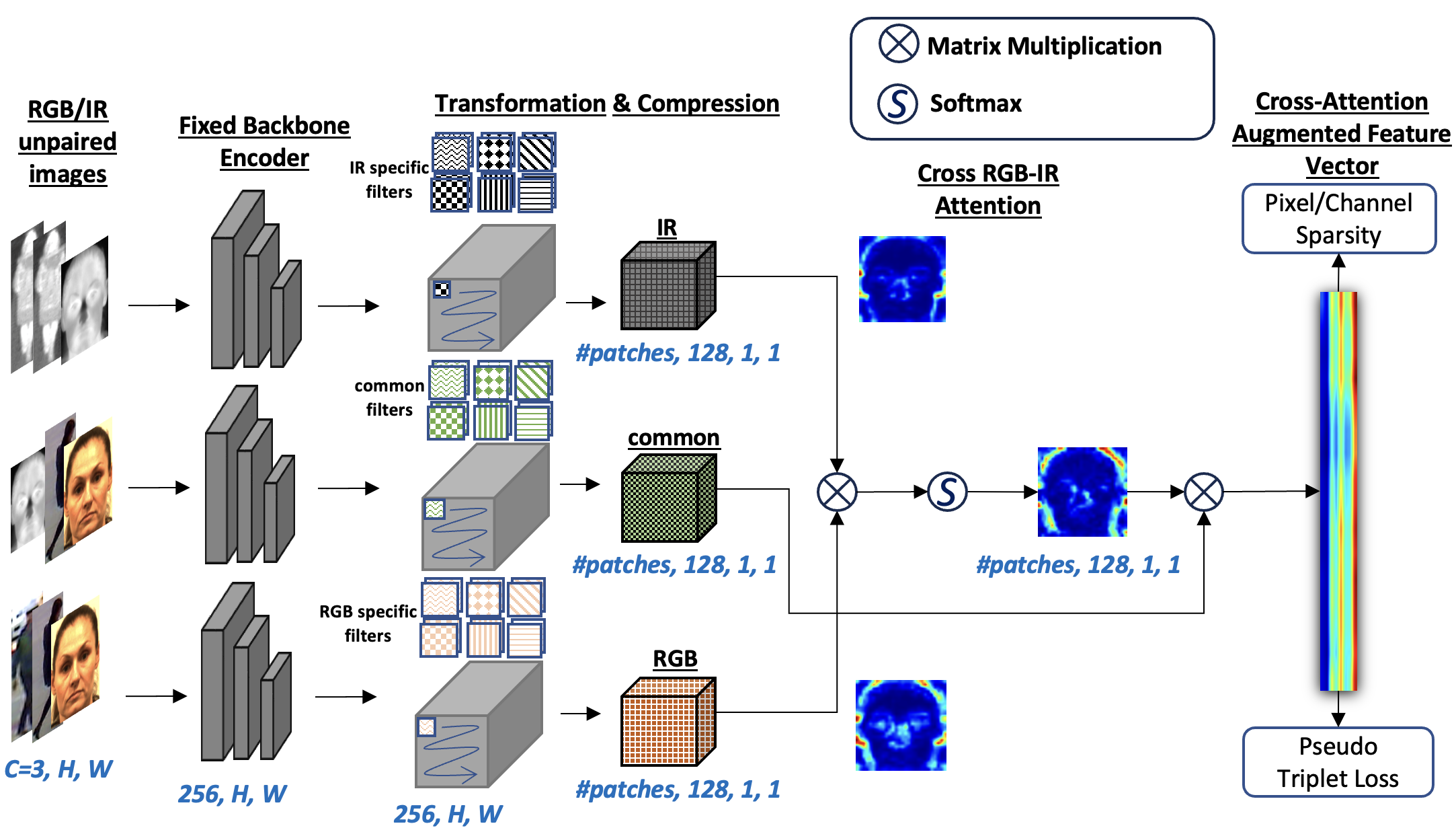}
    \caption{Our framework generates IR-specific, RGB-specific and a common feature representations from a shared truncated VGG backbone encoder and is optimized using our PTL voting scheme and pixel-channel sparsity term. IR-specific and RGB-specific representations are first used to compute the cross-spectral attention. Then, both RGB and IR (i.e., gallery and probe) common representations are both multiplied by the cross-spectrum attention to emphasize mutually beneficial characteristics. The corresponding dimensions are denoted in \textcolor{cyan}{\textbf{blue}} where $H$ is the height, $W$ is the width, $C$ is the number of input channels and \#$patches$ is the number of patches generated.}  
    \label{fig:methodology}
\end{figure*}

The proposed unsupervised cross-spectral framework (Fig.~\ref{fig:methodology}) consists of three main components: a cross-spectral attention network (CSAN), a pseudo triplet loss (PTL), and a pixel/channel sparsity (PCS) penalty. 
RGB and IR images (faces or whole body) are used to produce representations from three subspaces: RGB, IR, and common (i.e., both RGB and IR). Then, the cross-spectral attention is used to enhance the common representation by leveraging shared information between RGB and IR specific features. This representation is optimized by both PTL and PCS. Each component is discussed at length in this section. 


\begin{figure*}[ht]
    \centering
    \includegraphics[width=\textwidth]{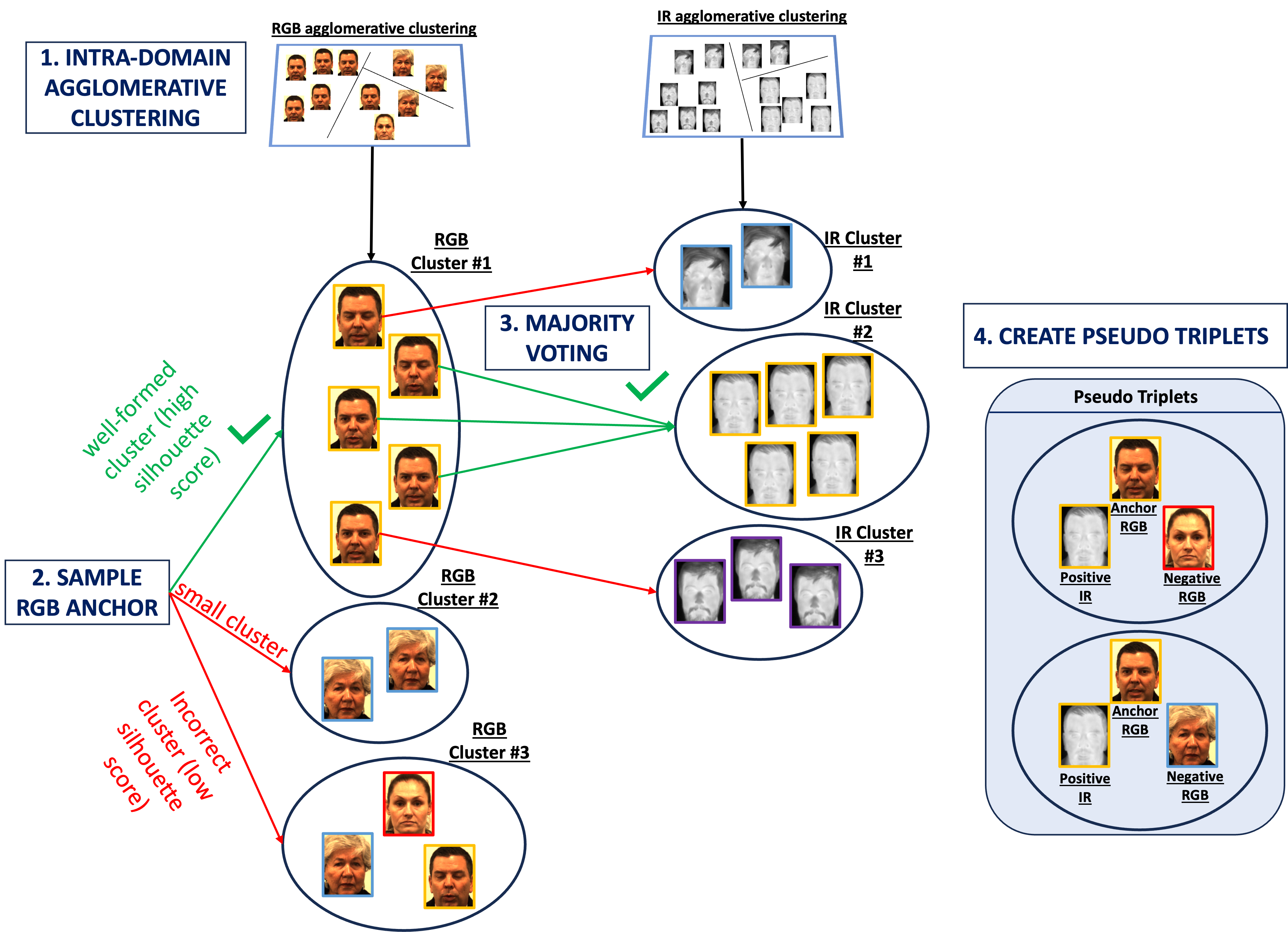}
    \caption{First, RGB and IR clusters are generated separately using agglomerative clustering. Next, an RGB cluster with a high cluster quality score is randomly sampled. Each sample of the selected RGB cluster votes for the nearest (in similarity) IR cluster. Lastly, pseudo anchor and positive samples are mined from newly associated RGB and IR cluster, and negative samples are mined from the nearest RGB samples.}
    \label{fig:triplet}
\end{figure*}

\subsection{Notation and Preliminaries}
\label{sec:preliminary}
We denote the unlabeled RGB and IR data as $X_{RGB}=\{x^{rgb}_1,\dots,x^{rgb}_n\}$ and $X_{IR}=\{x^{ir}_1,\dots,x^{ir}_n\}$ and the corresponding pseudo/cluster labels as 
$Y_{RGB}=\{y^{rgb}_1,\dots,y^{rgb}_n\}$ and $Y_{IR}=\{y^{ir}_1,\dots,y^{ir}_n\}$. 
Note that supervision via ground truth labels is not used throughout the training process. The goal of the framework is to learn a mapping that is agnostic to spectra by discriminatively clustering identities across RGB and IR domains in an unsupervised manner.  

\subsection{Intra-Domain Agglomerative Clustering}
As no labeled samples from either spectra are available for training, we generate pseudo-labels using agglomerative clustering~\cite{lin2019bottom,nikhal2021unsupervised, nikhal2022multi} for each domain.
Initially, every image is considered its own cluster. The cluster representations (i.e., centers) are stored in a memory bank matrix $M$ and optimized using the cross-entropy loss similar to ~\cite{nikhal2021unsupervised, nikhal2022multi}. Subsequently, the most similar clusters are merged at each iteration, where the Euclidean distance of the closest pair of features from two cluster quantifies the degree of cluster similarity. Given the pair-wise distance measures between every cluster pair, a fraction of the most similar clusters are merged.  
This is continued until obtaining the desired number of clusters. 
Since our cross-spectral framework intelligently samples pairs of RGB and IR clusters after agglomerative clustering, we do not require the desired number of clusters to equal the number of true identities for intra-domain agglomerative clustering.  Instead, our framework depends on having a sufficient number of high-quality---unknowingly correct but similar representations across both spectra---cluster pairs to help bridge domain gaps with pseudo-labels.

\subsection{Cross-Spectral Attention Network (CSAN)}
To generate feature representations that are domain agnostic, it is vital to leverage knowledge from both spectra. 
Previous studies~\cite{agw} argue that very low-level (first few blocks) features in the network contain domain specific information that should be adapted to address the domain gap. However, in this work, we use a fixed network with \textit{intermediate} representations and show that it is possible to learn a domain mapping leveraging the original, well-formed generic low-level features. Intermediate
features have been shown to better transfer between spectral domains~\cite{fondje2020cross}.
We use a truncated VGG16~\cite{vgg} network as a backbone,  
to acquire an intermediate feature representation of the original image.  The backbone can be denoted as $F_{VGG} = VGG_\ell(x_i)$ 
where $\ell$ represents the output from the $\ell^{th}$ block.
Then, this intermediate representation is adapted by our attention-network to generate a domain invariant feature representation. 

To learn this mapping, we leverage three representation subspaces, namely common, IR-specific and RGB-specific subspaces. Each subspace is generated by a $1\times1$ convolutional layer projecting from higher number of channels to fewer channels (e.g., $256\rightarrow128$). The intermediate representation is fed to the common, IR- and RGB-specific transformations, which are defined by
\begin{equation}
    F_{u} = tanh(W_u * F_{VGG})~\text{for } u \in \text{\{RGB, IR, common\}},
\end{equation}
where $W_u$ is a low-dimensional transformation parameterized by $1\times1$ kernel followed by hyperbolic activation functions that scale pre-activations to $[-1,1]$. 
Then, the cross-spectral attention is calculated using the scaled matrix product between the IR- and RGB-specific representation:

\begin{equation}
    A_{cross} = \log \left \{ softmax\left(\alpha F_{rgb}F_{ir}^{T}\right)\right \}
    \label{eq:crossattention}
\end{equation}
where $\alpha$ is a multiplier factor set to $1/\sqrt{dim}$ representing the relative importance of attention, $dim$ represents the channel dimension. We employ this scaled multiplicative attention similar to \cite{vaswani2017attention}, as for larger dimensions, the dot product becomes large, thereby having small gradients.

Finally, the output---cross attention augmented feature map---is the matrix product between cross-spectral attention (Eq.~\ref{eq:crossattention}) and the common representation:  
\begin{equation}
F_{out} = A_{cross}F_{common}.
\end{equation}

This facilitates learning of a low-dimensional transform that lies at the union of the two domain specific manifolds, leading to convergence with high accuracy.  
Note that if the image is RGB, no gradient is calculated for the IR subspace, and similarly for the IR image, no gradient is calculated for the RGB subspace. This ensures that the domain specific transformation learn domain specific filter information.

\subsection{Pseudo Triplet Loss with Offline Cross-Spectral Voting (PTL)}

To optimize the network, we leverage a triplet loss~\cite{schroff2015facenet} with cross-spectral vote sampling. 
Mathematically, for some distance on the feature space, the triplet loss can be defined as:

\resizebox{0.9\columnwidth}{!}{
\begin{minipage}{\columnwidth}
\begin{equation}
    L_{PTL} = max( d(F_{rgb}^{anchor}, F_{ir}^{positive}) - d(F_{rgb}^{anchor}, F_{rgb}^{negative}) + \beta, 0)
    \label{eq:ptl}
\end{equation}
\end{minipage}
}
where $d(\cdot)$ is the euclidean distance metric and $\beta$ is the margin. 
Given the RGB and IR pseudo-labels generated by agglomerative clustering, we need to bridge the spectral gap by incorporating the unknown cross-spectral pairs in the triplet loss. For this, we randomly sample a RGB cluster $i$ to act as the anchor, provided it has low intra-cluster distance and high inter-cluster distance, similar to the Silhoutte coefficient~\cite{silhoutte}.
Using this anchor, we generate a partial distance matrix computed between the cluster samples of the cluster $i$, and all other IR samples. Next, a voting scheme is used, where each RGB representation votes to be clustered with the IR cluster using the partial distance matrix. The majority IR cluster label is selected to be matched with this RGB cluster.
This is considered to be the positive IR sample for the triplet loss. The negative RGB is found using the hard negative mining scheme, where we find the closest RGB cluster that does not belong to $i$. Should the RGB cluster not receive a majority of votes, an alternative anchor RGB cluster is randomly selected for consideration. Fig.~\ref{fig:triplet} shows the overall voting sampling process.
This is repeated after every epoch such that the improved network can incorporate new clusters.\\
FaceVeri and ReID often consist of limited sized datasets. Hence, by using a triplet loss, various permutations can be created (anchor-positive and anchor-negative pairs) of the cross-spectral pairs, for each anchor point. This effectively increases the dataset by considering the relative relationships between identities. In addition, as our cross-spectral attention framework is very lightweight, that is, a parameter-efficient network having fewer number of trainable parameters, we are less prone to over-fitting and only a small amount of cross-spectral pairs are required to bridge the spectral gap and observe significant performance gain. Such parameter-efficient design choices are common in ReID (such as OSNet~\cite{zhou2019omni}.
Fig.~\ref{fig:noise} shows the number of clusters (incorrect and correct) and effect on rank-1 accuracy performance on the RegDB dataset. 

\subsection{Pixel-Channel Sparsity Representation Learning (PCS)}
While supervised learning has leveraged large-scale networks to learn from a tremendous amount of data,
unsupervised learning is more prone
to over-fitting and optimizing the network to incorrect minima due to the lack of annotations.
Regularizing using sparsity~\cite{liu2015sparse} and pruning~\cite{han2015learning} have decreased the over-fitting phenomena to some level, but in this case of learning with limited number of parameters and clustered data, further introducing pruning and sparsity in the network only limits the networks training capability. 
Hence, we utilize sparsity in the feature representation in the spatial and filter-wise manner to penalize unimportant filter responses. 
\begin{equation}
    L_{PCS}(F_{out}) = \sum_{c=1}^C \sum_{h=1}^H \sum_{w=1}^W ||F_{out}||_1 
    \label{eq:pcs}
\end{equation}
where $H$ is the height, $W$ is the width, $C$ is the number of channels, and $||\cdot||_1$ is the Manhattan/L1-norm. This works in conjunction with our cross attention, where our cross attention 
propagates important cross-spectral information, the sparsity further constraints it to generate a better network feature representation. This corroborates the study in SSL~\cite{wen2016learning}, where limiting the number of filters and filter shapes produced better network structure. However, in this case, we enforce sparsity on the representation itself. 

The overall objective function can be denoted as:
\begin{equation}
\label{eq:total}
L_{total} = L_{PTL} + \lambda \cdot L_{PCS}
\end{equation}
where $\lambda$ is a weighting hyper-parameter for the sparsity. An ablation study for $\lambda$ is presented in Fig.~\ref{fig:lambda_regdb} and Fig.~\ref{fig:lambda_arl}.
In addition to performance gain, PCS also encourages energy savings and faster inference. This is because many hardware accelerators~\cite{chen2016eyeriss, kim2017zena, zhu2020efficient} are able to perform efficient zero-activation skipping~\cite{9458578}. 

\begin{algorithm}[!h]
	\caption{Workflow Pseudocode} 
    \label{algo:workflow}
    \hspace*{\algorithmicindent} \textbf{Input:} Unlabeled data  $X_{RGB}=\{x^{rgb}_1,\dots,x^{rgb}_n\}$ \& 
    \hspace*{\algorithmicindent} $X_{IR}=\{x^{ir}_1,\dots,x^{ir}_n\}$ \\
    \hspace*{\algorithmicindent}  \textbf{Pseudo Labels:} $Y_{RGB}=\{y^{rgb}_1,\dots,y^{rgb}_n\}=\{1,\dots,n\}$ \hspace*{\algorithmicindent} \& $Y_{IR}=\{y^{ir}_1,\dots,y^{ir}_n\}=\{1,\dots,n\}$ \\
    
    \hspace*{\algorithmicindent} \textbf{Outputs:} Domain Agnostic Mapping $\phi(x;\theta)$\\
	\hspace*{\algorithmicindent} \textbf{Hyperparameters:} $\lambda=1e^{-3} (1e^{-5} \text{ for ARL}), \chi=0.6$\\
	\begin{algorithmic}[1]
    
    	
	    \State \textbf{\textcolor{orange}{// Intra-domain clustering and filtering}}
    	\State $Y_{RGB} = agglomerative\_clustering(X_{RGB})$ ~\cite{nikhal2021unsupervised}
    	\State $Y_{IR} = agglomerative\_clustering(X_{IR})$ ~\cite{nikhal2021unsupervised}
    	
	    \For {$n=1,2,\dots,N_{RGB}$} 
	    
	    \State  $S(F^n_{RGB}) = \frac{inter(F^n_{RGB}) - intra(F^n_{RGB})}{max(inter, intra)}$,
	    \If{$S(F^n_{RGB}) <= \chi$} 
	    \State $remove(X^n_{RGB}, Y^n_{RGB})$
	    \EndIf
	    \EndFor
	    
	    \For {$n=1,2,\dots,N_{IR}$} 
	    \State  $S(F^n_{IR}) = \frac{inter(F^n_{IR}) - intra(F^n_{IR})}{max(inter, intra)}$,
	    \If{$S(F^n_{IR}) <= \chi$} 
	    \State $remove(X^n_{IR}, Y^n_{IR})$
	    \EndIf
\EndFor
    	\textcolor{orange}{$\blacksquare$}
        \State
    	\State $N_{RGB} =$ number of unique RGB clusters
    	\State $N_{IR} =$ number of unique IR clusters

		\For {$epoch=1,2,\dots, 20$}
		    \State \textbf{\textcolor{orange}{// PTL cross-voting}}
		    
		    \State $F_{RGB} = \phi(X_{RGB};\theta)$ 
		    \State $F_{IR} = \phi(X_{IR};\theta)$ 
		    
		    \For {$n=1,2,\dots,N_{RGB}$} 
		        \State $D_{partial} = cdist(F^n_{RGB}, F_{IR})$
		        \State winner, count = $counter(argmin(D_{partial}))$ 
		        \If{$count >= 50\%$ of winner}  
		            \State $Y^{winner}_{IR} = Y^n_{RGB}$ 
		        \Else{}
		        \State $remove(X^n_{RGB}, Y^n_{RGB})$ 
		        
    		    \EndIf
                 \EndFor    	\textcolor{orange}{$\blacksquare$}

    		    \State  \textbf{\textcolor{orange}{ // Train}}
        		\For {$i=1,2,\dots$} 
        		    
        		    \State $L_{total} = L_{PTL} + \lambda \cdot L_{PCS}$
        		    \State $\theta = \theta - \eta \nabla_\theta L_{total}$ 
    		\EndFor\textcolor{orange}{$\blacksquare$}
    		    
		    \State
		    
		    \EndFor

	\end{algorithmic} 
\end{algorithm}

\subsection{Algorithmic Overview}
Algorithm~\ref{algo:workflow} presents an overview of our cross-modal learning framework. We start with unlabeled data in both spectral modalities, and randomly assign pseudo/cluster labels to each image in the dataset. Hyperparameters $\lambda$ denotes the weighting term for PCS and $\chi$ is the silhouette score threshold empirically determined to be 0.6. As mentioned in Section~\ref{sec:preliminary},  we perform intra-domain agglomerative clustering (Line 2, 3) to get our initial RGB and IR clusters. Next, we use the silhouette score to filter (Lines 4 to 15) poorly formed clusters in both domains:
\begin{equation}
    S(c_i) = \frac{inter(c_i) - intra(c_i)}{max(inter, intra)},
\end{equation}
where $inter(c_i)$ is the average nearest neighbour cluster distance and $intra(c_i)$ is the average intra-cluster distance for cluster sample $c_i$.
PTL iterates over each filtered RGB cluster, and finds the closest IR matches for each sample in this RGB cluster using the distance matrix between features. If a majority of these IR matches come from a single IR cluster, then we associate these two RGB and IR clusters together. Otherwise, this RGB cluster is not considered for the current training epoch (Line 19 to 31). Finally, we use these selected samples with the objective functions to perform a gradient descent step (Line 33-36). We continue this until a drop in validation performance. Code will be made publicly available.

\section{Experiments and Results}
\label{sec:experiments}
This section presents results on two public face and person ReID datasets to conduct comprehensive experiments using the proposed framework. The evaluation metrics are defined in Section~\ref{sec:metrics} and details of the implementation are outlined in Section~\ref{sec:implementation}. Many qualitative and quantitative experiments are presented, in addition to ablation studies pertaining to the different parameters in the framework. 
\subsection{Datasets}

\textbf{ARL Visible-Thermal Face Dataset:}
ARL-VTF~\cite{arlvtf} consists of over 500,000 thermal and visible images of 395 unique subjects (295 training, 100 testing). 
The dataset comprises two 100 subject gallery sets: G\_VB0- includes subject without glasses and G\_VB0+ include only 30 subjects wearing glasses. 
The probe sets are grouped into baseline, expression, and eye-wear conditions. Specifically, there are three different probe sets for baseline thermal imagery: P\_TB0, P\_TB-, and P\_TB+. The suffix `0' indicates imagery of subjects without glasses,`-' specifies the imagery of subjects who have glasses but were not wearing them, and `+' denotes the imagery from subjects that have glasses and had their glasses on. 
Additionally, there are two different probe sets for thermal imagery with varying expressions: P\_TE0 and P\_TE-. \\
\textbf{RegDB:}
The RegDB dataset~\cite{regdb} contains 412 persons, acquired using a RGB and thermal-IR camera. Using dual camera acquisition system, a total of 4,120 paired visible-IR frames are captured. Each identity has 10 RGB and thermal-IR images. Following the standard protocol, the data is equally split into the training and testing set that is randomly sampled, and is repeated over 10 trains to obtain an average of the systems performance. For the testing phase, both RGB-IR and IR-RGB matching modes are available.

\setlength{\tabcolsep}{4pt}
\begin{table*}[tb!]
\centering
\caption{Results on the RegDB ReID dataset (Infrared to Visible). * denotes not specifically developed for cross-spectral ReID. \textbf{\textcolor{blue}{Blue}} denotes best unsupervised performance, \textbf{\textcolor{cyan}{cyan}} denotes best supervised performance scores.}
\label{table:regdb}
\resizebox{1.3\columnwidth}{!}{

\begin{tabular}{lllllll}
\hline\noalign{\smallskip}
Method & Venue & Supervision & Rank-1 & Rank-10 & Rank-20 & mAP \\
\noalign{\smallskip}
\hline
\noalign{\smallskip}
HOG*~\cite{dalal2005histograms} & CVPR05 & ImageNet & 13.49 & 33.22 & 43.66 & 10.31 \\
LOMO*~\cite{lomo} & CVPR15 & ImageNet & 1.05 & 3.66 & 5.62 & 2.16 \\
CycleGAN*~\cite{zhu2017unpaired} & ICCV17 & ImageNet & 4.73 & 14.81 & 21.16 & 4.86 \\
HHL*~\cite{Zhong_2018_ECCV} & ECCV18 & ImageNet & 4.61 & 11.48 & 16.53 & 6.22 \\
ECN*~\cite{zhong2019invariance} & CVPR19 & ImageNet & 2.17 & 8.38 & 12.55 & 2.90 \\
H2H~\cite{homhet} & TIPS21 & Market1501 & 14.11 & 31.85 & 40.04 & 12.29\\
ADCA~\cite{yang2022augmented} & MM22 & ImageNet & 68.48 & 83.21 & 88.00 & 63.81\\
ACCL~\cite{wu2023unsupervised} & CVPR23 & ImageNet & 69.85  & - & - & \textbf{\textcolor{blue}{65.17}} \\
\hline
CSAN & - & ImageNet & 34.56 & 51.40 & 58.60 & 25.35 \\
CSAN+PTL & - & ImageNet & 76.84 & 90.38 & 93.39 & 54.76\\
CSAN+PTL+PCS & - & ImageNet & \textbf{\textcolor{blue}{78.59}} & \textbf{\textcolor{blue}{91.65}} & \textbf{\textcolor{blue}{94.02}} & 56.02 \\
\hline
One Stream~\cite{wu2017rgb} & ICCV17 & Full & 13.11 & 32.98 & - & 14.02 \\
Two Stream~\cite{wu2017rgb} & ICCV17 & Full & 12.43 & 30.36 & - & 13.42 \\
Zero Padding~\cite{wu2017rgb} & ICCV17 & Full & 17.75 & 34.21 & - & 18.90\\
TONE~\cite{tone} & AAAI18 & Full & 16.87 & 34.03 & 44.10 & 14.92 \\
BDTR~\cite{ye2018visible} & IJCAI18& Full & 33.47 & 58.42 & - & 31.83\\
MAC~\cite{mac} & MM19 & Full & 36.43 & 62.36 & 71.63 & 37.03 \\
DFE~\cite{hao2019dual} & MM19 & Full & 67.99 & 85.56 & 91.41 & 66.70 \\
JSIA~\cite{wang2020crossmodality} & AAAI20 & Full & 48.10 & - & - & 48.90 \\
DDAG~\cite{eccv20ddag} & ECCV20 & Full & 68.06 & 85.15 & 90.31 & 61.80\\
Xmodal~\cite{xmodality} & AAAI20 & Full & 62.21 & 83.13 & \textbf{\textcolor{cyan}{91.72}} & 60.18 \\
Hi-CMD~\cite{Choi_2020_CVPR} & CVPR20 & Full & 70.93 & \textbf{\textcolor{cyan}{86.39}} & - & 66.04 \\
cm-SSFT~\cite{cmssft} & CVPR20 & Full & 71.00 & - & - & 71.70 \\
cmAlign~\cite{learningbualigning} & ICCV21 & Full & 56.30 & - & - & 53.40 \\
AGW~\cite{agw} & TPAMI21 & Full & 70.05 & - & - & 66.37 \\
VSD~\cite{Tian_2021_CVPR} & CVPR21 & Full & 71.8 & - & - & 70.10 \\
TNL~\cite{Yang_2022_CVPR} & CVPR22 & Full & \textbf{\textcolor{cyan}{81.97}} & - & - & \textbf{\textcolor{cyan}{73.78}} \\

\hline
\end{tabular}
}
\end{table*}
\setlength{\tabcolsep}{1.0pt}

\subsection{Evaluation metrics}
\label{sec:metrics}
Performance on the RegDB dataset is measured using the rank-k (for k=1, 10, 20) retrieval accuracy, derived from the cumulative matching characteristic (CMC) curve. In addition, we also report mAP (mean Average Precision) scores derived from the precision-recall curve, which reflects on the separability among retrievals. 
Performance on the ARL-VTF dataset is measured by using the Area Under the Curve (AUC) metric, and also the True Accept Rate (TAR) at False Acceptance Rate (FAR) at $1\%$ and $5\%$. The equal error rate, where the FAR and False Rejection Rate (FRR) are the same, is used to measure the overall robustness level of the biometric system. 
\subsection{Implementation Details}
\label{sec:implementation}
The backbone encoder is a pre-trained VGG16 network, truncated at block3-conv3 layer to generate an intermediate representation. This layer was empirically chosen for faster convergence to bridge the spectral gap. Each of the RGB, IR and common transformation is a convolutional layer with a $1 \times 1$ kernel size and $128$ output channel dimensions. The images are resized to $128 \times 64$ for the RegDB dataset and $224 \times 224$ for the ARL-VTF dataset. To eliminate possible erroneous clusters, we discard extremely small/large clusters and silhoutte scores of less than $0.6$. The training process is continued until drop in validation performance. The proposed framework is implemented in PyTorch and trained on one NVIDIA RTX 2080 GPU with 11GB video memory. The batch size is set to 32 and the learning rate is set to 1e-4 with RMSProp optimizer.

\begin{table*}[tb]
    \caption{Verification performance comparisons for the baseline conditions 
    \textbf{\textcolor{blue}{Blue}} denotes best unsupervised performance, \textbf{\textcolor{cyan}{cyan}} denotes best supervised performance scores.}
    \centering
    \resizebox{1.3\columnwidth}{!}{
    \begin{tabular}{llllllllll}
    \toprule 
    \multicolumn{1}{c}{} & \multicolumn{1}{l}{} & \multicolumn{4}{c}{Gallery G\_VB0-} & \multicolumn{4}{c}{Gallery G\_VB0+} \\
    \cmidrule(lr){3-6} \cmidrule(lr){7-10}
    \multicolumn{1}{l}{Probes} & \multicolumn{1}{l}{Method} & AUC & EER &  FAR=1\% & FAR=5\% & AUC & EER &  FAR=1\% & FAR=5\% \\
    
    \midrule  {\multirow{4}{*}{P\_TB0}}
    & Raw & 61.37 & 43.36 & 3.13  & 11.28 & 62.83 & 42.37 & 4.19 & 13.29 \\
    & Pix2Pix~\cite{pix2pix2017}  & 71.12 & 33.80  &  6.95 & 21.28 & 75.22 & 30.42 & 8.28 & 27.63 \\
    & GANVFS~\cite{ganvfs} & 97.94 & 8.14 & 75.00  & 88.93 & 98.58 & 6.94 & 79.09 & 91.04 \\
    & SAGAN~\cite{sagan}  & 99.28 & 3.97 & 87.95  & 96.66 & 99.49 & 3.38 & 90.52 & 97.81 \\
    & LG-GAN~\cite{lggan}  & 96.96 & 5.94 & - & - & - & - & - & - \\
    & RST~\cite{rst}  & 99.76 & 2.30 & 96.84 & 98.43 & 99.87 & 1.84 & 97.29 & 98.80 \\
    & PDT~\cite{pdt}  & 99.95 & 1.13 & 98.57 & 100.00 & 99.95 & 1.14 & 98.57 & 100.0 \\
    & DPIT~\cite{dpit}  & \textbf{\textcolor{cyan}{99.99}} & \textbf{\textcolor{cyan}{0.15}} & \textbf{\textcolor{cyan}{100.00}} & \textbf{\textcolor{cyan}{100.00}} & \textbf{\textcolor{cyan}{100.00}} & \textbf{\textcolor{cyan}{0.12}} & \textbf{\textcolor{cyan}{100.00}} & \textbf{\textcolor{cyan}{100.00}} \\
    \cdashline{2-10}
    & CSAN  & 97.58 & 7.58 & 83.60 & 93.32 & 98.48 & 5.83 & 83.61 & 93.19 \\
    & CSAN+PTL  & 99.55 & \textbf{\textcolor{blue}{2.00}} & 96.78 & \textbf{\textcolor{blue}{98.43}} & 99.58 & 1.82 & 97.25 & \textbf{\textcolor{blue}{98.43}} \\
    & CSAN+PTL+PCS  & \textbf{\textcolor{blue}{99.56}} & \textbf{\textcolor{blue}{2.00}} & \textbf{\textcolor{blue}{97.14}} & \textbf{\textcolor{blue}{98.43}} & \textbf{\textcolor{blue}{99.59}} & \textbf{\textcolor{blue}{1.80}} & \textbf{\textcolor{blue}{97.55}} & \textbf{\textcolor{blue}{98.43}} \\
    \cdashline{1-10}
    {\multirow{1}{*}{P\_VB0}}
    & GT Vis-to-Vis & 99.99 & 0.23 & 99.79  & 99.95 & 99.99  & 0.24 & 99.86 & 100.00 \\
    \cmidrule(lr){1-10}
    
    {\multirow{4}{*}{P\_TB-}}
    & Raw & 61.14 & 41.64 & 2.77  & 16.11 & 57.61 & 44.73 & 1.38 & 6.11 \\
    & Pix2Pix~\cite{pix2pix2017} & 68.77 & 38.02  &  6.69 & 20.28 & 52.11 & 48.88 & 2.22 & 4.66 \\
    & GANVFS~\cite{ganvfs} & 99.36 & 3.77 & 84.88  & 97.66 & 87.34 & 18.66 & 7,00 & 29.66 \\
    & SAGAN~\cite{sagan} & 99.63 & 2.66 & 91.55  & 98.88 & 89.24 & 19.49 & 16.33 & 41.22 \\
    & RST~\cite{rst} & 99.83 & 1.95 & 96.00 & 99.48 & 99.03 & 4.79 & 85.56 & 95.86 \\
    & PDT~\cite{pdt}  & 99.96 & 1.18 & 98.67 & 100.00 & 99.94 & 1.33 & 98.67 & 100.00 \\
    & DPIT~\cite{dpit} & \textbf{\textcolor{cyan}{100.00}} & \textbf{\textcolor{cyan}{00.0}} & \textbf{\textcolor{cyan}{100.00}} & \textbf{\textcolor{cyan}{100.00}} & \textbf{\textcolor{cyan}{97.97}} & \textbf{\textcolor{cyan}{0.66}} & \textbf{\textcolor{cyan}{100.00}} & \textbf{\textcolor{cyan}{100.00}} \\
    
    \cdashline{2-10}
    & CSAN  & 98.30 & 5.30 & 78.58 & 94.17 & 93.26 & 13.31 & 61.44 & 78.20 \\
    & CSAN+PTL  & \textbf{\textcolor{blue}{99.88}} & \textbf{\textcolor{blue}{1.63}} & \textbf{\textcolor{blue}{97.34}} & \textbf{\textcolor{blue}{99.73}} & \textbf{\textcolor{blue}{98.35}} & \textbf{\textcolor{blue}{5.93}} & \textbf{\textcolor{blue}{76.64}} & \textbf{\textcolor{blue}{93.60}} \\
    & CSAN+PTL+PCS  & 99.87 & \textbf{\textcolor{blue}{1.63}} & 96.73 & 99.72 & 98.28 & 6.03 & 76.62 & 93.16 \\
    \cdashline{1-10}
     {\multirow{1}{*}{P\_VB-}}
    & GT Vis-to-Vis & 100.00 & 0.00 & 100.00  & 100.00 & 99.06  & 4.33 & 89.66 & 96.22 \\
    
    \cmidrule(lr){1-10}
    
    {\multirow{4}{*}{P\_TB+}}
    & Raw & 59.52 & 42.60 & 4.66  & 6.00 & 78.26 & 29.77 & 3.88 & 21.33 \\
    & Pix2Pix~\cite{pix2pix2017} & 59.68 & 41.72  &  3.33 & 3.33 & 67.08 & 36.44 & 2.68 & 11.11 \\
    & GANVFS~\cite{ganvfs}  & 87.61 & 20.16 & 20.55  & 44.66 & 96.82 & 8.66 & 46.77 & 83.00 \\
    & SAGAN~\cite{sagan} & 91.11 & 17.43 & 22.33  & 55.66 & 97.96 & 7.21 & 60.11 & 88.70 \\
    & RST~\cite{rst}  & 99.28 & 5.32 & 89.21 & 94.79 & 99.97 & 0.73 & 99.47 & 100.00 \\
    & PDT~\cite{pdt}  & 99.48 & 4.11 & 89.33 & 97.33 & 99.60 & 4.00 & 90.00 & 97.33 \\
    & DPIT~\cite{dpit}  & \textbf{\textcolor{cyan}{99.91}} & \textbf{\textcolor{cyan}{1.94}} & \textbf{\textcolor{cyan}{96.84}} & \textbf{\textcolor{cyan}{100.00}} & \textbf{\textcolor{cyan}{100.00}} & \textbf{\textcolor{cyan}{0.32}} & \textbf{\textcolor{cyan}{100.00}} & \textbf{\textcolor{cyan}{100.00}} \\
    \cdashline{2-10}
    & CSAN  & 95.99 & 11.03 & 51.77 & 75.89 & 98.04 & 6.56 & 76.75 & 90.01 \\
    & CSAN+PTL  & 98.56 & 6.61 & 85.76 & 90.90 & \textbf{\textcolor{blue}{99.95}} & \textbf{\textcolor{blue}{6.32}} & \textbf{\textcolor{blue}{100.00}} & \textbf{\textcolor{blue}{100.00}} \\
    & CSAN+PTL+PCS  & \textbf{\textcolor{blue}{98.69}} & \textbf{\textcolor{blue}{6.35}} & \textbf{\textcolor{blue}{86.77}} & \textbf{\textcolor{blue}{91.51}} & \textbf{\textcolor{blue}{99.95}} & 7.00 & 99.91 & \textbf{\textcolor{blue}{100.00}} \\
    \cdashline{1-10}
     {\multirow{1}{*}{P\_VB+}}
    & GT Vis-to-Vis & 99.62 & 2.70  & 93.22 & 98.51  & 99.92 & 1.44 & 98.55 & 99.66 \\
    \bottomrule
    \end{tabular} 
    }
    \label{tab:arl}
\end{table*}

\begin{table}[tb]
    \caption{Face verification performance comparisons for expression conditions. }
    \centering
    \resizebox{1.0\columnwidth}{!}{
    \begin{tabular}{llllllllll}
    \toprule 
    \multicolumn{1}{c}{} & \multicolumn{1}{l}{} & \multicolumn{4}{c}{Gallery G\_VB0-} & \multicolumn{4}{c}{Gallery G\_VB0+} \\
    \cmidrule(lr){3-6} \cmidrule(lr){7-10}
    \multicolumn{1}{l}{Probes} & \multicolumn{1}{l}{Method} & AUC & EER &  FAR=1\% & FAR=5\% & AUC & EER &  FAR=1\% & FAR=5\% \\
    \cmidrule(lr){1-10}

    {\multirow{4}{*}{P\_TE0}}
    & Raw & 61.40 & 41.96 & 3.40  & 12.18 & 62.50 & 41.38 & 4.60 & 13.25 \\
    & Pix2Pix~\cite{pix2pix2017}  & 69.10 & 35.98  &  7.01 & 16.44 & 73.97 & 31.87 & 7.93 & 19.60 \\
    & GANVFS~\cite{ganvfs}  & 96.81 & 10.51 & 70.41  & 84.00 & 97.73 & 8.90 & 74.20 & 86.80 \\
    & SAGAN~\cite{sagan}  & 98.46 & 6.44 & 81.11  & 92.49 & 98.89 & 5.60 & 84.23 & 93.94 \\
    & RST~\cite{rst}  & 98.95 & 3.61 & 92.61 & 96.88 & 99.01 & 3.57 & 92.69 & 96.93 \\
    & PDT~\cite{pdt}  & 99.90 & 1.72 & 97.43 & 99.77 & 99.90 & 1.72 & 97.43 & 99.77 \\
     & DPIT~\cite{dpit}  & \textbf{\textcolor{cyan}{99.79}} & \textbf{\textcolor{cyan}{2.39}} & \textbf{\textcolor{cyan}{96.49}} & \textbf{\textcolor{cyan}{98.31}} & \textbf{\textcolor{cyan}{99.70}} & \textbf{\textcolor{cyan}{2.33}} & \textbf{\textcolor{cyan}{96.52}} & \textbf{\textcolor{cyan}{98.29}} \\
    
    \cdashline{2-10}
    & CSAN  & 96.10 & 9.28 & 79.47 & 87.74 & 96.13 & 9.23 & 80.49 & 87.86 \\
    & CSAN+PTL  & \textbf{\textcolor{blue}{97.50}} & 6.79 & 86.87 & 92.06 & \textbf{\textcolor{blue}{97.55}} & \textbf{\textcolor{blue}{6.66}} & \textbf{\textcolor{blue}{89.40}} & \textbf{\textcolor{blue}{92.62}} \\
    & CSAN+PTL+PCS  & 97.40 & \textbf{\textcolor{blue}{6.77}} & \textbf{\textcolor{blue}{88.29}} & \textbf{\textcolor{blue}{92.28}} & 97.41 & 6.86 & 89.06 & 92.40 \\
    \cdashline{1-10}
     {\multirow{1}{*}{P\_VE0}}
    & GT Vis-to-Vis & 99.98 & 5.38  & 99.65 & 99.92  & 99.98 & 0.45 & 99.73 & 99.96 \\
    \cmidrule(lr){1-10}
    
    {\multirow{4}{*}{P\_TE-}}
    & Raw & 63.26 & 42.34 & 4.66  & 16.28 & 59.33 & 43.17 & 2.04 & 8.00 \\
    & Pix2Pix~\cite{pix2pix2017}  & 68.78 & 36.24  &  7.75 & 18.06 & 51.05 & 49.11 & 2.26 & 4.95 \\
    & GANVFS~\cite{ganvfs} & 98.66 & 5.93 & 73.17  & 92.82 & 83.68 & 22.41 & 6.77 & 22.13 \\
    & SAGAN~\cite{sagan} & 99.30 & 3.84 & 82.55  & 97.44 & 86.12 & 21.68 & 9.88 & 31.62 \\
    & RST~\cite{rst}  & 99.83 & 2.27 & 95.66 & 99.48 & 99.48 & 3.05 & 89.45 & 98.07 \\
    & PDT~\cite{pdt}  & 99.95 & 0.93 & 99.07 & 100.00 & 99.90 & 1.73 & 97.87 & 100.00 \\
    & DPIT~\cite{dpit}  & \textbf{\textcolor{cyan}{99.88}} & \textbf{\textcolor{cyan}{0.81}} & \textbf{\textcolor{cyan}{99.47}} & \textbf{\textcolor{cyan}{99.87}} & \textbf{\textcolor{cyan}{99.77}} & \textbf{\textcolor{cyan}{2.92}} & \textbf{\textcolor{cyan}{95.33}} & \textbf{\textcolor{cyan}{98.87}} \\
    \cdashline{2-10}
    & CSAN  & 98.93 & 5.60 & 86.87 & 93.90 & 97.88 & 7.53 & 69.46 & 86.67 \\
    & CSAN+PTL  & \textbf{\textcolor{blue}{99.81}} & \textbf{\textcolor{blue}{2.33}} & \textbf{\textcolor{blue}{94.13}} & \textbf{\textcolor{blue}{99.53}} & \textbf{\textcolor{blue}{98.88}} & \textbf{\textcolor{blue}{5.87}} & 78.30 & \textbf{\textcolor{blue}{93.12}} \\
    & CSAN+PTL+PCS  & 99.79 & 2.41 & 93.91 & \textbf{\textcolor{blue}{99.53}} & 98.79 & 6.34 & \textbf{\textcolor{blue}{78.42}} & 92.41 \\
    \cdashline{1-10}
     {\multirow{1}{*}{P\_VE-}}
    & GT Vis-to-Vis & 99.99 & 0.14  & 99.97 & 99.97  & 97.96 & 6.69 & 72.16 & 90.91 \\
    \cmidrule(lr){1-10}

    \end{tabular} 
    }
    \label{tab:arl2}
\end{table}

\begin{table}[h!]
    \caption{Face verification performance comparisons for pose conditions. }
    \centering
    \resizebox{1.0\columnwidth}{!}{
    \begin{tabular}{llllllllll}
    \toprule 
    \multicolumn{1}{c}{} & \multicolumn{1}{l}{} & \multicolumn{4}{c}{Gallery G\_VB0-} & \multicolumn{4}{c}{Gallery G\_VB0+} \\
    \cmidrule(lr){3-6} \cmidrule(lr){7-10}
    \multicolumn{1}{l}{Probes} & \multicolumn{1}{l}{Method} & AUC & EER &  FAR=1\% & FAR=5\% & AUC & EER &  FAR=1\% & FAR=5\% \\
    \cmidrule(lr){1-10}

    {\multirow{4}{*}{P\_TP0}}
    & Raw & 55.24 & 46.25 & 2.23 & 8.25 & 55.10 & 46.34 & 2.91 & 8.74 \\
    & Pix2Pix~\cite{pix2pix2017}  &  54.86 & 47.22 & 3.13 & 9.78 & 56.50 & 46.03 & 4.01 & 10.84 \\
    & GANVFS~\cite{ganvfs}  &  63.70 & 41.66 & 16.55 & 23.73 & 65.58 & 40.19 & 17.95 & 25.68 \\
    & SAGAN~\cite{sagan}  & 65.06 & 40.24 & 17.33 & 24.56 & 67.13 & 38.67 & 18.91 & 26.46 \\
    & RST~\cite{rst}  & 66.26 & 38.05 & 22.18 & 30.72 & 68.39 & 36.86 & 22.64 & 31.81 \\
    & PDT~\cite{pdt}  & 87.56 & 20.57 & 60.80 & 68.86 & 87.51 & 20.57 & 60.86 & 68.86 \\
    & DPIT~\cite{dpit}  & \textbf{\textcolor{cyan}{97.69}} & \textbf{\textcolor{cyan}{7.75}} & \textbf{\textcolor{cyan}{66.08}} & \textbf{\textcolor{cyan}{88.74}} & \textbf{\textcolor{cyan}{97.39}} & \textbf{\textcolor{cyan}{8.39}} & \textbf{\textcolor{cyan}{68.00}} & \textbf{\textcolor{cyan}{88.54}} \\
    \cdashline{2-10}
    & CSAN  & 60.89 & 43.80 & 10.76 & 19.83 & 65.27 & 38.11 & 7.56 & 17.60 \\
    & CSAN+PTL  & 61.40 & \textbf{\textcolor{blue}{43.40}} & 14.59 & \textbf{\textcolor{blue}{21.54}} & 62.73 & 42.60 & 14.97 & 22.25 \\
    & CSAN+PTL+PCS  & \textbf{\textcolor{blue}{61.56}} & 43.46 & \textbf{\textcolor{blue}{14.71}} & \textbf{\textcolor{blue}{21.54}} & \textbf{\textcolor{blue}{63.06}} & \textbf{\textcolor{blue}{42.28}} & \textbf{\textcolor{blue}{15.10}} & \textbf{\textcolor{blue}{22.31}} \\
    \cdashline{1-10}
     {\multirow{1}{*}{P\_VP0}}
    & GT Vis-to-Vis & 75.76 & 32.30 & 28.54 & 35.52 & 77.24 & 30.92 & 29.49 & 37.27 \\
    \cmidrule(lr){1-10}
    
    {\multirow{4}{*}{P\_TP-}}
    & Raw & 55.48 & 45.98 & 3.25 & 8.47 & 56.82 & 44.74 & 2.09 & 7.57 \\
    & Pix2Pix~\cite{pix2pix2017}  & 54.31 & 47.04 & 2.93 & 8.44 & 50.08 & 49.67 & 0.60 & 4.33 \\
    & GANVFS~\cite{ganvfs} & 65.79 & 40.35 & 17.84 & 25.48 & 59.51 & 44.04 & 4.29 & 15.47 \\
    & SAGAN~\cite{sagan} & 67.27 & 39.00 & 18.16 & 26.02 & 60.10 & 43.57 & 5.77 & 15.97 \\
    & RST~\cite{rst} & 68.24 & 37.60 & 23.09 & 33.54 & 63.29 & 41.79 & 18.79 & 27.93 \\
    & PDT~\cite{pdt} & 87.78 & 20.40 & 65.33 & 71.20 & 87.30 & 20.65 & 60.00 & 69.87 \\
    & DPIT~\cite{dpit}  & \textbf{\textcolor{cyan}{97.09}} & \textbf{\textcolor{cyan}{9.09}} & \textbf{\textcolor{cyan}{63.91}} & \textbf{\textcolor{cyan}{84.40}} & \textbf{\textcolor{cyan}{96.62}} & \textbf{\textcolor{cyan}{10.39}} & \textbf{\textcolor{cyan}{55.84}} & \textbf{\textcolor{cyan}{78.73}} \\
    \cdashline{2-10}
    & CSAN  & 58.50 & 45.41 & 3.21 & 11.20 & 54.02 & 48.20 & 2.32 & 8.82 \\
    & CSAN+PTL & \textbf{\textcolor{blue}{59.13}} & 44.54 & 13.25 & 20.27 & \textbf{\textcolor{blue}{55.47}} & 48.00 & \textbf{\textcolor{blue}{11.58}} & 18.01 \\ 
    & CSAN+PTL+PCS & 59.07 & \textbf{\textcolor{blue}{44.26}} & \textbf{\textcolor{blue}{13.76}} & \textbf{\textcolor{blue}{20.59}} & 55.17 & \textbf{\textcolor{blue}{47.47}} & 11.53 & \textbf{\textcolor{blue}{18.20}} \\
    \cdashline{1-10}
     {\multirow{1}{*}{P\_VP-}}
    & GT Vis-to-Vis & 75.59 & 33.37 & 29.37 & 36.64 & 69.62 & 37.61 & 19.36 & 28.11 \\
    \cmidrule(lr){1-10}

    \end{tabular} 
    }
    \label{tab:arlpose}
\end{table}

\subsection{Quantitative Analysis}
\textbf{RegDB:} Table~\ref{table:regdb} presents the results on the person re-identification task using the RegDB dataset. The first five entries correspond to the unsupervised and unsupervised domain adaptation learning (using a visible-only ReID dataset), while the rest (except our proposed work) utilize full label supervision to train their model. We present three settings of our proposed method, namely: CSAN, CSAN+PTL, and CSAN+PCS. CSAN is trained using our cross-spectral attention network with a traditional cross-entropy loss. CSAN + PTL combines our attention network with our offline-voting based triplet loss. CSAN + PTL + PCS denotes the addition of the pixel-channel sparsity term as denoted in Eq.~\ref{eq:total}.
Compared to the latest H2H model, we achieve substantial performance gain of 20.45\% in rank-1 accuracy and 13.06\% in mAP scores with the baseline. Using our PTL loss, a remarkable rank-1 accuracy performance of 76.84\% is noticed, which surpasses many recent SOTA supervised performances. This is possible because of the lightweight cross-spectral attention sub-network and a sampling that iteratively ensures only high confident clusters is passed through the network. Constraining the network with a PCS penalty, the rank-1 accuracy improves to 78.59\% and the mAP increases to 56.02\%. While H2H also compares with their re-ranking approach, re-ranking optimizes gallery-to-gallery similarity and is applied only during inference. We compare only feature extractors among recent methods. We suspect that using re-ranking techniques would enhance the discriminability of our image representations, since re-ranking methods~\cite{reranking} show consistent performance gain irrespective of feature extraction method used. As we do not utilize any re-ranking metric, it would be an unfair comparison with methods using re-ranking. Similarly, with AGW~\cite{agw}, we only wanted to compare independent methods and not combination of multiple recent methods. 
ADCA~\cite{yang2022augmented} and ACCL~\cite{} comes closest to our SOTA unsupervised performance, and has better mAP scores. However, ADCA depends on DBSCAN with predefined hyper-parameters, whereas our approach involves learning to generate pseudo-labels through clustering similar examples. This leads to generalized clustering performance across datasets, eliminating the need for fine-tuning clustering algorithms hyper-parameters.
While our scores do not compete with most recent SOTA supervised technique, we suspect it is because of the improved separability gained between classes using ground truth label supervision.\\ 
Table~\ref{table:regdb2} also shows additional results on the RegDB dataset from the Visible-to-Infrared setting, i.e., the query consists of RGB images and the gallery consists of IR images. The results are consistent with our Infrared-to-Visible setting, thereby showcasing our methods ability to bridge the spectral domain gap. Many methods in the literature do not specify the setting they are using, or do not report performance on both these settings as Infrared-to-Visible is most applicable to nighttime recognition.
 \\

\begin{figure*}[h!]
    \centering
    \includegraphics[width=1.5\columnwidth]{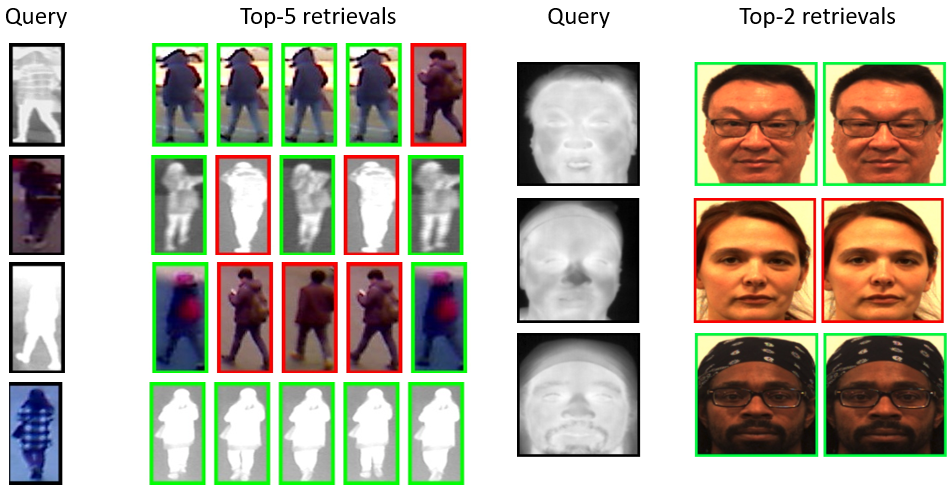}
    \caption{The top-5 and top-2 retrievals on the person and face biometric tasks. \textcolor{green}{Green} denotes correct retrieval while \textcolor{red}{red} denotes incorrect retrieval. }
    \label{fig:ranking}
\end{figure*}

\begin{figure*}[h]
     \centering
     \begin{subfigure}[b]{0.45\columnwidth}
         \centering
         \includegraphics[width=\columnwidth]{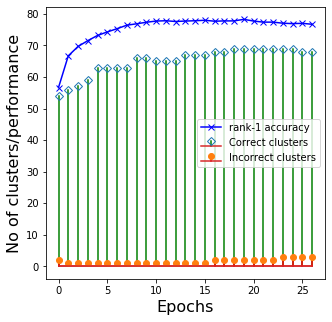}
         \caption{Performance trend (\textcolor{blue}{blue line}) after each epoch. \textcolor{teal}{Green} stems denote correct clusters, while \textcolor{orange}{orange} stems denotes incorrect clusters.}
         \label{fig:noise}
     \end{subfigure}
     ~
      \begin{subfigure}[b]{0.45\columnwidth}
         \centering
         \includegraphics[width=\columnwidth]{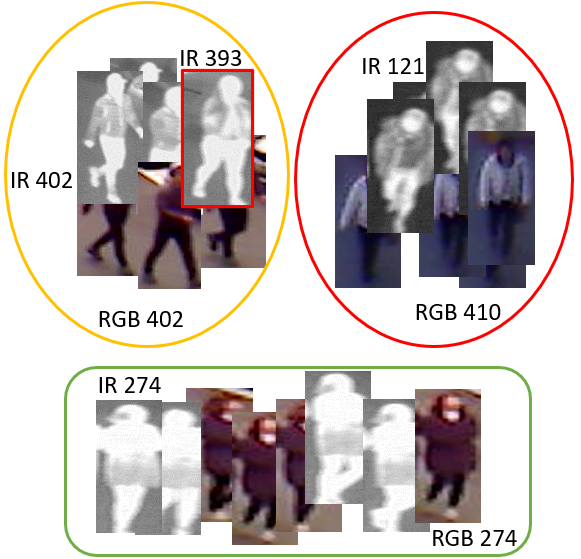}
         \caption{\textcolor{red}{Incorrect}, \textcolor{orange}{partially correct} and \textcolor{green}{correct} clusters formed in the set while training.}
         \label{fig:cluster_tsne}
     \end{subfigure}
     ~
     \begin{subfigure}[b]{0.47\columnwidth}
         \centering
         \includegraphics[width=\columnwidth]{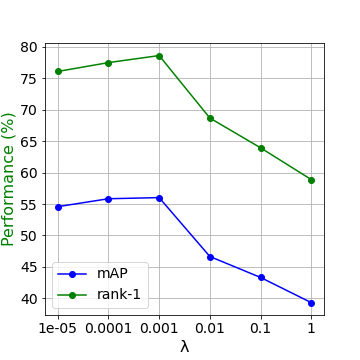}
         \caption{$\lambda$ values for RegDB.}
         \label{fig:lambda_regdb}
     \end{subfigure}
     \begin{subfigure}[b]{0.45\columnwidth}
         \centering
         \includegraphics[width=\columnwidth]{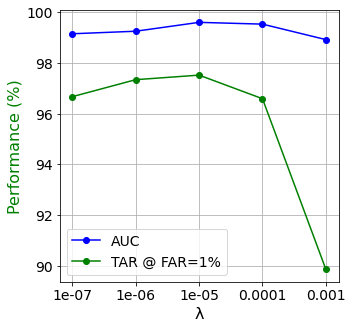}
         \caption{$\lambda$ values for ARL-VTF.}
         \label{fig:lambda_arl}
     \end{subfigure}
     \caption{Training performance trends (a), cluster formation analysis (b), and hyper-parameter analysis (c, d).}
     \label{fig:tsne_noise}
\end{figure*}

\textbf{ARL-VTF:}
Table~\ref{tab:arl} showcases baseline performance (i.e., frontal faces with neural expressions) on the FaceVeri task on the ARL-VTF dataset. The first row section uses probe set P\_TB0 with galleries G\_VB0- and G\_VB0+. All methods (except our proposed work) uses full ground truth label supervision for training. Compared to GAN methods like Pix2Pix, GANVFS and SAGAN, CSAN+PTL+PCS achieves improvement in TAR @ FAR=1\% by 89.83\%, 21.78\%, and 8.83\%, respectively on the P\_TB0 probe set. RST performs slightly better than CSAN+PTL, but with the addition of PCS, it sets a new unsupervised benchmark of 97.14\% TAR @ FAR=1\%. The recent DPIT~\cite{dpit} method achieves the best supervised performance on most conditions, but explicitly uses pose correction unlike other supervised methods. Similar performance trends are seen for different gallery sets and with the probe set P\_TB-. For the probe set (P\_TB+), CSAN+PTL+PCS achieves perfect face biometric performance with 100\% TAR with the VB0+ gallery set.
For the expression condition (P\_TE0 and P\_TE-) in Table~\ref{tab:arl2}, it again achieves impressive performance compared to GAN methods, and achieves close to SOTA supervised performance without using any label annotations.

Table~\ref{tab:arlpose} shows results on the ARL dataset on the pose condition. Obtaining pseudo labels amid large pose variations is an extremely challenging task. While we do not address the pose condition in this work, we present it here for completeness.  Our method achieves close to supervised performance in some metrics even without explicitly performing any pose correction. 






          



\subsection{Qualitative Analysis}

\textbf{Ranking Analysis:}
Figure~\ref{fig:ranking} shows the retrieval results on the RegDB and ARL-VTF dataset. For the RegDB dataset, we can see that retrievals with the highest similarity score yield correct ReID result, explaining the high rank-1 accuracy score while false positives in between correct retrievals hurts mAP scores. The proposed work does well in both RGB-IR and IR-RGB settings and is able to successfully ignore dominant RGB features such as color and texture. Row-3 has a very saturated IR heat signature, and thus results in incorrect retrievals.
For the ARL dataset, the framework is able to successfully retrieve correct identities across different expressions and eyewear (row 1) and head gear (row 3). Row 2 shows a false positive, where an identity with a similar face structure is retrieved.\\
\textbf{Training Cluster Analysis:}
Figure~\ref{fig:noise} shows the number of incorrect (\textcolor{orange}{orange stems}) and correct cross-spectral clusters (\textcolor{green}{green stems}) in the training set and the rank-1 accuracy (\textcolor{blue}{blue line}) through consecutive training epochs using PTL. As is evident, the performance improves because of the mostly correct associations in the training set and the number of correct clusters increase while reducing the number of incorrect clusters. Over-training and/or reducing the cluster quality thresholds in PTL introduces errors as is evident after epoch around 18. Figure~\ref{fig:cluster_tsne} shows an example of training clusters, where the orange cluster has the correct association between domains (IR 402 \& RGB 402), but has an IR identity 393 incorrectly clustered with IR 402 (because of the initial intra-domain agglomerative clustering). The red cluster has wrong RGB-IR associations because of the similarity between poses and low contrast in the RGB image. The green cluster shows perfect association.

\begin{figure}[h]
     \centering
     \begin{subfigure}[]{0.8\columnwidth}
         \centering
         \includegraphics[width=\columnwidth]{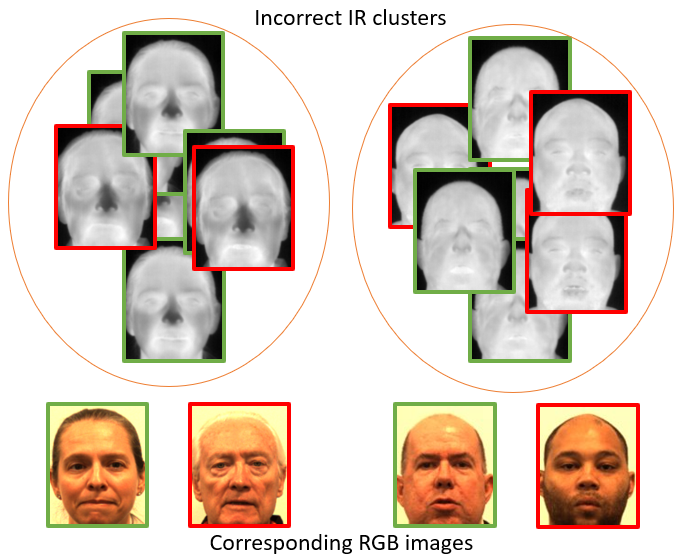}
    \caption{Incorrect IR-IR clustering}
    \label{fig:incorrect_ir}
     \end{subfigure}
     ~
     \begin{subfigure}[]{0.8\columnwidth}
         \centering
          \includegraphics[width=\columnwidth]{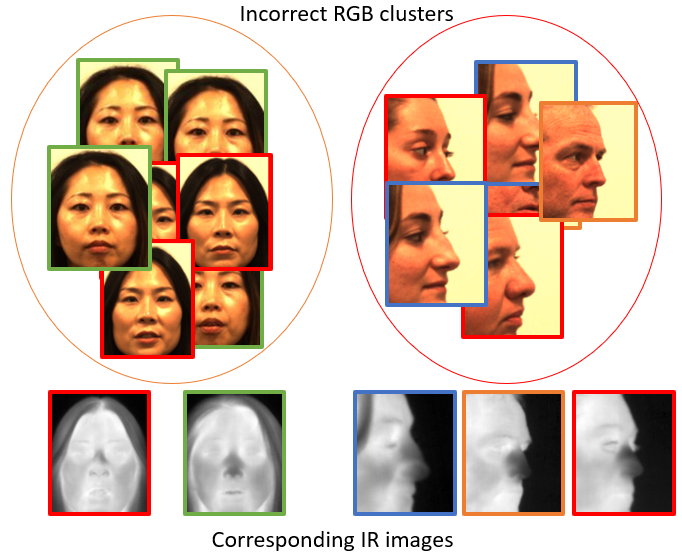}
    \caption{Incorrect RGB-RGB clustering}
    \label{fig:incorrect_rgb}
     \end{subfigure}

     \begin{subfigure}[]{0.6\columnwidth}

         \centering
          \includegraphics[width=\columnwidth]{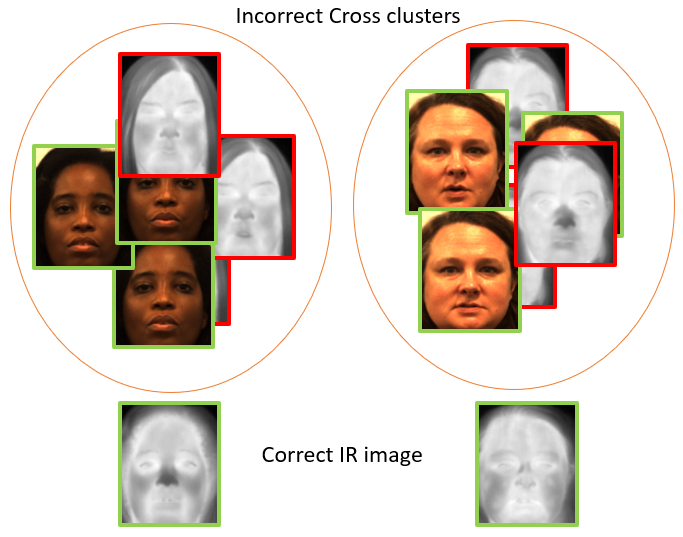}
    \caption{Incorrect RGB-IR clustering}
    \label{fig:incorrect_cross}
         \end{subfigure}

     \caption{Clustering performance }
     \label{fig:correctincorrectposes}
\end{figure}

Additionally, Figure~\ref{fig:incorrect_ir} and Figure~\ref{fig:incorrect_rgb} shows errors in the intra-domain clustering process that impacts performance.  In Figure~\ref{fig:incorrect_ir}, we can see that our method clusters IR images that has a similar face structure and heat signature. However, the \textcolor{red}{red} outlined images do not belong to that identity, and we can see a clear difference in the corresponding RGB images. This is difficult to capture because of the lack of texture and color details in IR images. 
Likewise, in Figure~\ref{fig:incorrect_rgb}, similar looking identities are incorrectly clustered together (left image) but can see a distinct face structure in thermal images. Hence, complementary information between RGB and IR modalities are used to produce our cross-attention. Some RGB clusters suffer from cluster mode collapse because of lack of pose correction, where a single RGB pose (right image) is incorrectly clustered ignoring the identity's features. PTL reduces these errors by filtering out clusters with a low silhouette score. 
Figure~\ref{fig:incorrect_cross} shows incorrect RGB-IR cross clusters, where the RGB and IR face structure matches closely, but is not the correct IR cluster to be combined.  Our PTL loss tries to mitigate this effect by choosing high confident clusters in which each image votes to be clustered, instead of a single cluster centroid.

\begin{figure}[h]
     \centering
     \begin{subfigure}[b]{0.8\columnwidth}
         \centering
         \includegraphics[width=\columnwidth]{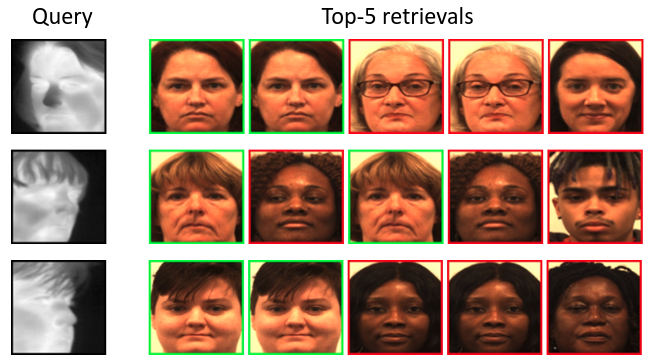}
         \caption{Correct retrievals with high variability in pose.
         }
         \label{fig:pose_a}
     \end{subfigure}
     ~
     \begin{subfigure}[b]{0.8\columnwidth}
         \centering
         \includegraphics[width=\columnwidth]{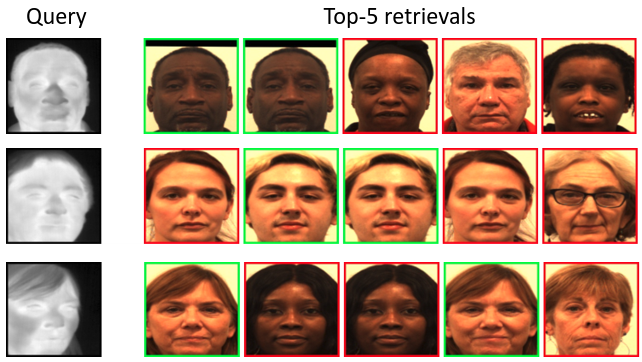}
         
         \caption{Correct retrievals with low variability in pose.
         }
         \label{fig:pose_b}
     \end{subfigure}

     \begin{subfigure}[ht]{0.8\columnwidth}

         \centering
         \includegraphics[width=\columnwidth]{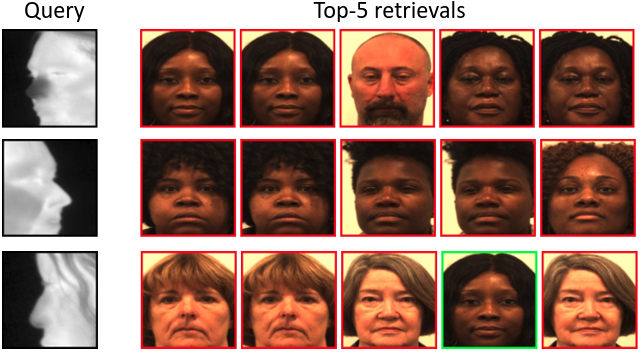}
          
         \caption{Incorrect retrievals with high variability in pose.
         }
         \label{fig:pose_c}
         \end{subfigure}

     \caption{Retrievals on the varying pose setting. }
     \label{fig:lowvariabilitypose}
\end{figure}

\textbf{Pose results:}
Figure~\ref{fig:lowvariabilitypose} shows the ranking results on the ARL-VTF dataset for the pose setting.  In Figure~\ref{fig:pose_a}, we see some positive results where a highly offset IR face image is used as a query, and our method is able to correctly retrieve the identity. Figure~\ref{fig:pose_b} shows a similar trend with a slightly offset face. However, as seen in Figure~\ref{fig:pose_c}, the method is not robust to large variations in pose, which explains the relatively low performance compared to baseline and expression conditions.

\subsection{Ablation Study}

\textbf{Compression:}
Table~\ref{tab:compression} studies the effect of varying compression in the feature transformation ($\tau_u$). The input feature size from the backbone is a 256-channel dimensional feature. Compressing it to a 128-channel dimensions achieves best rank-1 accuracy with less than 100,000 trainable parameters and feature vector size of 16,384. Compressing further impacts performance, but enables faster (and more scalable) search because of the reduction in feature size. We opted to use 128-channel dimensions for our experiments.

\begin{table}[tb]
    \centering
    \resizebox{0.6\columnwidth}{!}{%
    \begin{tabular}{ccccc}
            \hline

         Output Size &  Dim &  Parameters & rank-1 & mAP  \\
             \hline

         32 & 4,096 & 24,672 & 74.85 & 52.58\\
         64 & 8,192 & 49,344 & 75.87 & \textcolor{green}{55.04} \\
         128 & 16,384 & 98,688 & \textcolor{green}{76.84} & 54.76 \\
         256 & 32,768 & 197,376 & 76.35 & 54.47 \\
         512 & 65,536 & 394,752 & 75.97 & 54.05 \\
    \hline

    \end{tabular}%
    }
    \caption{The effect of compression on the RegDB dataset using PTL only. 
    Best performance is in \textcolor{green}{green}.}
    
    \label{tab:compression}
\end{table}

\textbf{$\lambda$ weighting:}
The amount of PCS (Eq.~\ref{eq:pcs}) to be enforced on the representation is added to the final objective and controlled by $\lambda$. Fig.~\ref{fig:lambda_regdb} shows the rank-1 accuracy and mAP scores for the RegDB dataset. As seen, the performance improves from $76.8\%$ to ~$78.6\%$ and $54.5\%$ to $56.1\%$, as $\lambda$ reaches 0.001. Adding too much sparsity restricts the capability of the network to produce a discriminative representation. 
Similarly, for the ARL dataset in Fig.~\ref{fig:lambda_arl}, performance improves upto ~$97.1\%$ and ~$99.56\%$ for $\lambda=1e-4$ and reduces as too much sparsity penalty is applied. The scales of $\lambda$ vary because of the different sizes of the feature representation.

\textbf{Visible to Infrared: } Table~\ref{table:regdb2} shows additional results on the RegDB dataset from the Visible to Infrared setting, i.e., the query consists of RGB images and the gallery consists of IR images. The results are consistent with our Infrared to Visible setting, thereby showcasing our methods ability to bridge the spectral domain gap. Many methods in the literature do not specify the setting they are using, or do not report performance on both these settings. Therefore, we only present our method in this table.

\setlength{\tabcolsep}{4pt}
\begin{table}
\centering
\caption{Results on the RegDB ReID dataset (Visible to Infrared).}
\label{table:regdb2}
\resizebox{\columnwidth}{!}{

\begin{tabular}{lllllll}
\hline\noalign{\smallskip}
Method & Venue & Supervision & Rank-1 & Rank-10 & Rank-20 & mAP\\
\noalign{\smallskip}
\hline
\noalign{\smallskip}

\hline
CSAN & - & ImageNet & 32.03 & 49.46 & 55.7 & 23.2 \\
CSAN+PTL & - & ImageNet & 73.44 & 87.62 & 90.92 & 50.83 \\
CSAN+PTL+PCS & - & ImageNet & \textbf{\textcolor{blue}{74.85}} & \textbf{\textcolor{blue}{89.32}} & \textbf{\textcolor{blue}{92.33}} & \textbf{\textcolor{blue}{52.75}} \\

\hline
\end{tabular}
}
\end{table}
\setlength{\tabcolsep}{1.0pt}

\setlength{\tabcolsep}{4pt}

\section{Conclusion}

This work assessed the problem of cross-spectral re-identification and verification without using any intra- and inter-domain annotations or associations by proposing a novel unsupervised cross-spectral attention framework. The framework used agglomerative clustering principles for intra-domain clustering and bridged the domain gap using (a) a cross-spectral attention network to leverage knowledge from both domains (b) a pseudo triplet loss that utilizes a novel sampling scheme and (c) a structurally consistent sparsity constraint to encourage distinct, and useful features. Although our work is one of the first to approach this problem in a completely unsupervised manner, we compared with recent SOTA semi-supervised and supervised techniques, and outperformed many methods in terms of multiple metric scores. We expect this work will set a new standard for unsupervised learning for cross-spectrum applications, and provide generalization on tasks beyond face and person biometrics.


\bibliographystyle{IEEEtran}
\bibliography{references}
\newpage

 





\end{document}